\documentclass{article}

\usepackage{microtype}
\usepackage{graphicx}
\usepackage{subfigure}
\usepackage{booktabs} 

\usepackage{hyperref}
\usepackage{fancyhdr}

\usepackage[accepted]{icml2024}

\usepackage{amsmath}
\usepackage{amssymb}
\usepackage{mathtools}
\usepackage{amsthm}

\usepackage[capitalize,noabbrev]{cleveref}

\theoremstyle{plain}

\theoremstyle{definition}

\theoremstyle{remark}

\newcommand{\eg}{\textit{e.g.,}~}

\usepackage{listings} 

\usepackage{algorithm}
\usepackage{algorithmic}

\definecolor{codegreen}{rgb}{0,0.6,0}
\definecolor{codegray}{rgb}{0.5,0.5,0.5}
\definecolor{codepurple}{rgb}{0.58,0,0.82}
\definecolor{backcolour}{rgb}{0.95,0.95,0.92}

\lstdefinestyle{mystyle}{
  commentstyle=\color{codegreen},
  keywordstyle=\color{magenta},
  numberstyle=\tiny\color{codegray},
  basicstyle=\ttfamily\footnotesize,
  stringstyle=\color{codepurple},
  breakatwhitespace=false,         
  breaklines=true,                 
  captionpos=b,                    
  keepspaces=true,                 
  numbers=left,                    
  numbersep=5pt,                  
  showspaces=false,                
  showstringspaces=false,
  showtabs=false,                  
}
\usepackage[most]{tcolorbox}

\newcommand{\paragraphspace}{\vspace{-3mm}} 

\icmltitlerunning{Automated Statistical Model Discovery with Language Models}

\begin{document}

\twocolumn[
\icmltitle{Automated Statistical Model Discovery with
Language Models}

\begin{icmlauthorlist}
\icmlauthor{Michael Y. Li}{yyy}
\icmlauthor{Emily B. Fox}{yyy,xxx,sss}
\icmlauthor{Noah D. Goodman}{yyy,zzz}
\end{icmlauthorlist}

\icmlaffiliation{yyy}{Department of Computer Science, Stanford University}
\icmlaffiliation{xxx}{Department of Statistics, Stanford University}
\icmlaffiliation{sss}{Chan Zuckerberg Biohub -- San Francisco}
\icmlaffiliation{zzz}{Department of Psychology, Stanford University}

\icmlcorrespondingauthor{Michael Y. Li}{michaelyli@stanford.edu}
\icmlkeywords{Machine Learning, ICML}

\vskip 0.3in
]



\printAffiliationsAndNotice{}  

\begin{abstract}
Statistical model discovery is a challenging search over a vast space of models subject to domain-specific constraints. 
Efficiently searching over this space requires expertise in modeling and the problem domain. 
Motivated by the domain knowledge and programming capabilities of large language models (LMs), we introduce a method for \textit{language model driven automated statistical model discovery}.
We cast our automated procedure within the principled framework of Box’s Loop: the LM iterates between proposing statistical models represented as \textit{probabilistic programs}, acting as a modeler, and critiquing those models, acting as a domain expert.
By leveraging LMs, we do not have to define a domain-specific language of models or design a handcrafted search procedure, which are key restrictions of previous systems.
We evaluate our method in three settings in probabilistic modeling: searching within a restricted space of models, searching over an open-ended space, and improving expert models under natural language constraints (\eg this model should be interpretable to an ecologist).
Our method identifies models on par with human expert designed models and extends classic models in interpretable ways.
Our results highlight the promise of LM-driven model discovery.
\end{abstract}

\section{Introduction}
Modeling, or generating a parsimonious but explanatory representation of a complex system, is at the heart of scientific discovery. 
Model discovery is challenging because it involves searching over a vast space of candidate models subject to domain-specific constraints (\eg find the best model that remains interpretable to domain experts). 
Efficiently searching over the space requires extensive human expertise:
modelers need broad knowledge of different modeling approaches and must work closely with domain experts to adapt these approaches to a given problem domain. 
As a concrete example, consider modeling blood-glucose dynamics in Type 1 diabetes (T1D) patients; accurately modeling these dynamics can enable better insulin regulation and reduce complications from the disease.  
To model these dynamics, modelers need to understand biomedical models that capture blood-glucose dynamics in idealized, lab settings but they also need to understand techniques for adapting these models to handle real data with noise and missingness \citep{pmlr-v126-miller20a}.
Domain experts play a crucial role in this process: modelers must work closely with clinicians to ensure that the model is consistent with human physiology. 
As this example illustrates, model discovery can require significant human expertise.
Automating this process could accelerate and democratize scientific discovery.
\begin{figure*}[ht!]
\centering
\includegraphics[width=0.95\textwidth]{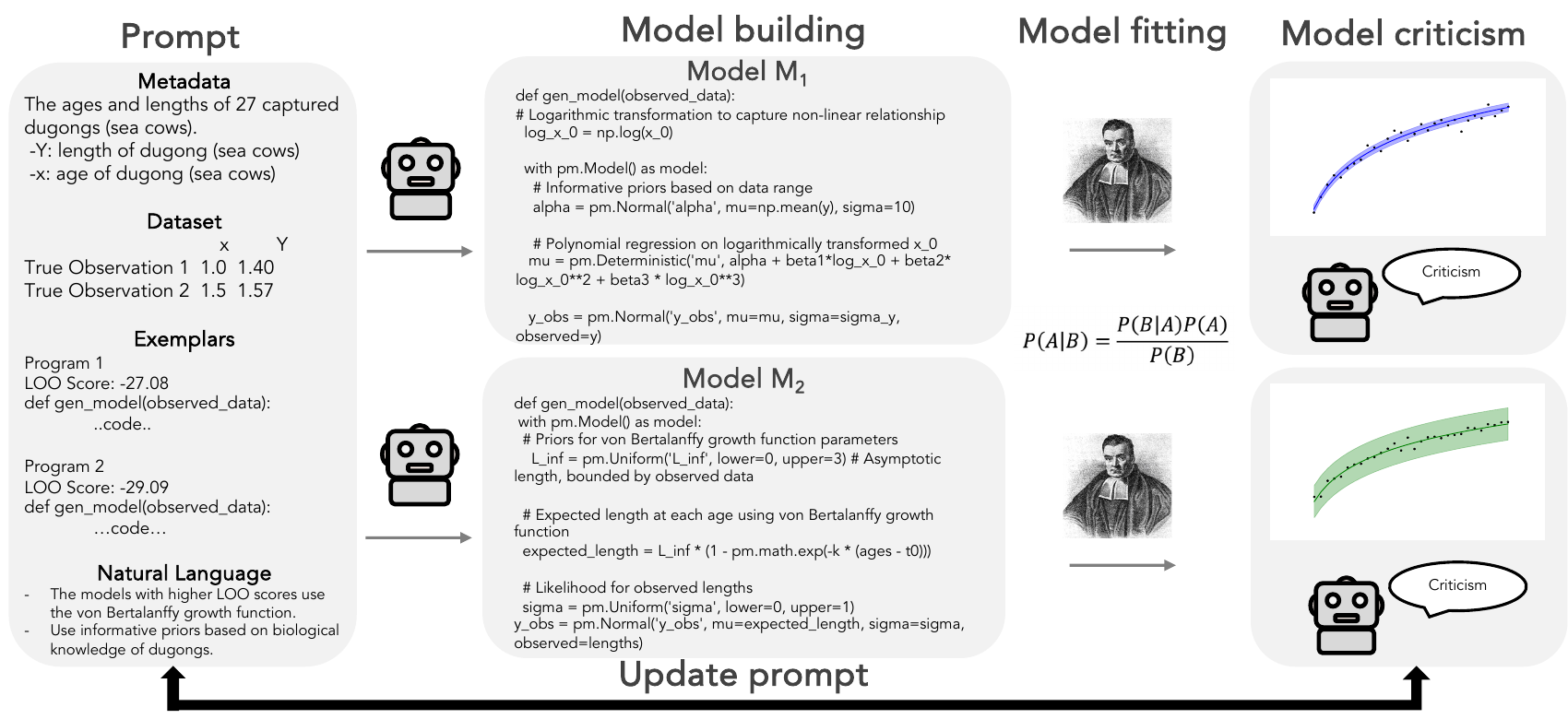}    
\caption{\textbf{Language model driven automated model discovery (\texttt{BoxLM}).}
1) The prompt for the LM contains the dataset in visual and/or textual form, dataset metadata (\eg dataset description), the code for previous probabilistic programs, and natural language feedback.
2) Given this, the proposal LM proposes new models expressed as \textit{probabilistic programs}. 
3) To fit these programs in a generic way, we leverage probabilistic programming languages and obtain scores and posterior predictive samples.
4) After we fit models, we compute the posterior predictive mean and variance.
We provide these statistics to a critic LM which produces natural language feedback to guide the next round of model building.
5) We propagate the best programs, their posterior predictive means and variances, and natural language feedback forward by updating the prompt. }
\label{fig:box_loop}    
\end{figure*}

Automated model discovery is not a new ambition.
Previous systems have been successfully deployed for discovering physical laws \citep{Bongard2007AutomatedRE, mckinney_phys_review, LINKA2023134}, reverse-engineering non-linear dynamical systems \citep{schmid_lipson_science}, nonparametric regression \citep{pmlr-v28-duvenaud13} and unsupervised learning \citep{Grosse2014ModelSI}.
However, in these systems, a human expert had to carefully design a domain specific language (DSL) of models and specify a hand-crafted search procedure for composing models in that DSL.  
For example, in the Automatic
Statistician \citep{pmlr-v28-duvenaud13, Lloyd2014AutomaticCA}, a system for {nonparametric regression and time series modeling, human experts defined a DSL of Gaussian process kernels and a search procedure that composes kernels via addition and multiplication.  
Defining the DSL and the operators for composing models requires significant modeling expertise.
These systems also compromised flexibility for automation: rather than choosing a model class best-suited for a problem, experts chose models that compose conveniently. 
This breaks the core principle of separation of modeling from inference \citep{vandemeent2021introduction}.

As we saw above, there are two key roles in the model discovery pipeline: the domain expert and the modeler.
We hypothesize that LMs can supplement these roles and reduce the human expertise required in model discovery. 
Our hypothesis is motivated by recent work exploring LM capabilities.
First, LMs have been successfully applied to domains including law \citep{Bommarito2022GPTTT}, medicine \citep{lee_nejm}, and mathematics \citep{wu2023empirical}. 
This suggests that LMs have broad domain knowledge which may enable them to supplement the domain expert. 
Second, LMs can reliably write code \citep{Rozire2023CodeLO, Chen2021EvaluatingLL} which means we do not require a human expert to define a DSL.
Instead, we can search over a more open-ended space of models, provided they can be expressed in a generic programming language like Python. 
Third, LMs have strong inductive reasoning capabilities (\eg they can generate hypotheses from limited data) \citep{wang2023hypothesis, qiu2024phenomenal}.
We hypothesize these capabilities may enable them to reason about data \citep{zhong2023goal}.
These capabilities, in addition to their knowledge of various modeling approaches, may enable the LM to supplement the modeler.

Leveraging LMs for automated model discovery is enticing at a conceptual level.
However, in order to deliver on this promising idea, we need to make several important design choices. These design choices should exploit the strengths of LMs but remain grounded in principled statistical modeling. 
First, we need a generic and flexible \textit{representation} of a statistical model that can be expressed programmatically and is amenable to automated inference (\eg model fitting).
Second, we need a method for \textit{guiding LM proposals through natural language}.
\begin{algorithm}
\caption{Automated Model Discovery with LMs}\label{alg:cap}
\begin{algorithmic}[!ht]
\INPUT dataset $\mathcal{D}$, number of rounds $T$, $k$ number of exemplars, $m$ number of proposals per round,  (optional) warm-start example $z_0$, function for scoring a program \texttt{score}, (optional) function for producing natural language feedback \texttt{criticize} 
\STATE $\mathcal{Z} \gets \emptyset$
\WHILE{$t<T$}
\STATE $\{z^t_i\}_{i=1}^m \sim q_{\text{LM}}(\cdot | \mathcal{Z}, z_0, h^{t}, \mathcal{D})$
\STATE $\{s_i\}_{i=1}^m \gets \texttt{score-all}(\texttt{score}, \{z^t_i\}_{i=1}^m, \mathcal{D})$
\STATE $\mathcal{Z} \gets \texttt{select-exemplars}(k, \{z^t_i\}_{i=1}^m, \{s_i\}_{i=1}^m)$
\STATE $h^{t+1} \gets \texttt{criticize}(\{z^t_i\}_{i=1}^m, \{s_i\}_{i=1}^m, h^t)$
\ENDWHILE
\label{alg:main_algo}
\end{algorithmic}
\end{algorithm}

To fulfill these requirements, we draw on research in probabilistic programming and inductive reasoning with LMs.
In particular, we introduce the following method: LMs propose statistical models expressed as \textit{probabilistic programs}, given a dataset and some metadata (\eg dataset description).
We then fit these models using generic probabilistic inference techniques, compute statistics assessing the model fit, summarize findings from these statistics in natural language, and select exemplar models to guide the next round of proposals (Figure~\ref{fig:box_loop}).
Our approach is connected to recent work on hypothesis search and inductive reasoning with LMs, but targets a fundamentally different problem \citep{wang2023hypothesis, qiu2024phenomenal}.
While we also leverage the inductive reasoning capabilities of LMs, our focus is on \textit{statistical modeling of noisy real-world data}, while previous work focused on learning deterministic input-output rules that can be implemented with standard Python programs.

We evaluate our method in three settings that cover common use cases in probabilistic modeling: searching \textit{within} a DSL, searching over an \textit{open-ended} space of probabilistic models, and improving expert models subject to modeling constraints expressed in natural language. 
In the first use case,  we illustrate that our LM system is effective even in a DSL and matches the performance of the 
{Automatic} Statistician \citep{pmlr-v28-duvenaud13}.  
We then consider the more general setting of automatically constructing
probabilistic models for real world data; crucially, we do not require a user to define a DSL and this generality is enabled by our choice to use probabilistic programs. 
Our method identifies probabilistic programs that match the performance of human expert written programs.  
In the third setting, we use LMs to improve classic models and illustrate a compelling advantage of using LMs for model discovery: given that certain modeling constraints can be difficult to express formally but easy to express in natural language (\eg this model should be more physical), we use natural language to guide LMs towards models that balance interpretability and flexibility.
\section{Automated Box's Loop with Language Models}
We begin with a brief background on the probabilistic modeling paradigm.
We then formally introduce our problem setting and describe our approach. 
For an overview, see Figure~\ref{fig:box_loop} and Algorithm~\ref{alg:main_algo}. 

\subsection{Background}
\label{sec:background}
In probabilistic modeling, our goal is to describe a dataset in terms of unobserved, latent structure.
We describe a dataset through a \textit{probabilistic model} which 
is a joint distribution $p(x, z| \eta)$; 
here $x=x_{1:N}$ denotes $N$ observed data points, $z = z_{1:M}$ denotes $M$ latent variables, and $\eta$ corresponds to non-random quantities in the model \citep{blei_build}.
After specifying a probabilistic model, we fit the model to observed data.
Fitting a model involves \textit{inference}, or conditioning on observed data and computing the \textit{posterior distribution} $p(z|x)$.
After fitting a model, we perform \textit{model criticism}.  
In the model criticism step, we evaluate the model by interrogating the posterior. 
In this work, our model criticism is inspired by a common model criticism technique known as a \textit{posterior predictive check}: to identify discrepancies, samples are drawn from the posterior predictive distribution and statistics of these samples are compared against those of the observed data \citep{Gelman2013BayesianDA}.
Model building and model criticism typically take place over multiple iterations in an iterative process known as \textit{Box's Loop} \citep{Box1962AUM}.

There are many different representations of a probabilistic model. 
Since our goal is automated model discovery, we use \textit{probabilistic programs}.
Probabilistic programming languages provide ways to flexibly represent probabilistic models as programs and support generic inference methods for any arbitrary program \citep{wood-aistats-2014, Goodman2008ChurchAL, vandemeent2021introduction}. 
In the context of automated model discovery, these are highly desirable properties, since LMs can write code reliably and other representations of probabilistic models can require custom inference methods.
Our approach of using LMs to generate probabilistic programs is related to the approach taken by \citet{Wong2023FromWM} but is motivated by a different problem. 
Our emphasis is on using LMs as a tool for developing statistical models of real-world datasets, while their focus was on integrating symbolic methods with LMs. 
\label{method}
\subsection{Problem formulation}
Our framework is motivated by recent work in inductive reasoning with LMs \citep{qiu2024phenomenal, wang2023hypothesis}, integrating tools with LMs \citep{gal_pal}, and driving LMs via linguistic feedback \citep{shinn2023reflexion}. 

At a high level, we consider a method for \textit{learning probabilistic models from data} that involves two steps: a \textit{model building}
step and a \textit{criticism} step\footnote{To avoid overloading the word ``model'', we will refer to language models as LMs.}.
Crucially, by learning a model, 
we mean searching over a space of model structures and \textit{not just learning the parameters of some fixed model class}. 
In each step, we leverage LMs.
In the proposal step, a \textit{proposal LM} proposes probabilistic programs for a dataset.
We then fit these probabilistic programs and evaluate them. 
In the criticism step, we provide a \textit{critic LM} with programs and statistics assessing model fit (\eg model criticism statistics) and ask the critic LM to provide feedback to guide the next round of proposals. 

We start with a dataset $\mathcal{D} = \{\mathbf{x_i}, y_i\}_{i=1}^n$. 
Here $\mathbf{x_i} \in \mathbb{R}^{d}$ are fixed $d$-dimensional input values (\eg features) and $y_i \in \mathbb{R}$ are the observations. 
Let $\Sigma$ be the vocabulary of the LM.
For each dataset, we have an associated metadata set $\mathcal{C} \in \Sigma^{d+1}
$, which consists of natural language descriptions of $\mathcal{D}$ (\eg animal ages vs length) and natural language descriptions of each feature in $\mathcal{D}$ (\eg length of animal). 
Context informs how human modelers leverage prior knowledge; for example, if a modeler knows their dataset consists of monthly carbon dioxide measurements over a fifty-year time span, they will choose a model that can capture periodicity and a linear trend.
Our goal is to find a probabilistic program $z \sim \Sigma^{*}$ that maximizes some notion of quality, which we take here to be either the log marginal likelihood or expected log predictive density (ELPD) estimated via cross validation (LOO) \citep{10.1007/s11222-016-9696-4}.
\subsection{Approach}
\paragraph{Model Building Step}
In the \textit{model building step}, we automatically generate probabilistic programs for modeling a dataset given information about the dataset and previously proposed programs.
In particular, to generate candidates for round $t$, we sample $m$ probabilistic programs $z^t_i$ from the proposal LM, $q_{\text{LM}}(\cdot)$.
In our experiments, we use GPT-4 V \citep{Achiam2023GPT4TR} (\texttt{gpt4-11-06-preview}), which has multimodal capabilities. 
We leverage \textit{in-context learning}, or LM's ability to learn from examples in a prompt, to guide the LM's proposals based on high-scoring programs in the past \citep{brown_few_shot}.
Specifically, $q_{\text{LM}}$ 
``conditions'' on  $h^{t} \in \Sigma^{*}$, a natural language instruction synthesizing previous modeling approaches and suggesting new approaches, $k$ exemplars $\{z_1, \ldots z_k\}$, and a visual or textual representation of $\mathcal{D}$. 
Optionally, 
$q_{\text{LM}}$
also conditions on a warm-start expert program $z_0$: 
\begin{align*}    
    z_i^t \sim q_{\text{LM}}(\cdot | z_0, z_1, \ldots, z_k, h^{t-1}, \mathcal{D}).
\end{align*}
We run this at a temperature of 0.7.
Chain-of-thought reasoning, or generating intermediate reasoning steps, improves the performance of LMs \citep{wei2022chain, kojima2022large}.
Motivated by this, we instruct $q_{\text{LM}}$ 
to reflect on the properties of the dataset or 
plot of the data, sketch a high-level modeling approach, state the hypotheses that it will address before writing a program, and add comments to code that address specific hypotheses.

To create exemplars $z_1,\dots,z_k$ for round $t$ for $q_{\text{LM}}$, 
we choose the best $k$ programs among the $m$ proposed programs in round $t-1$.
\paragraphspace{}
\paragraph{Model Fitting Step}
In the \textit{model fitting step}, we fit a probabilistic program to data.
This requires us to perform (approximate) inference for generic probabilistic models. 
To accomplish this, we leverage \texttt{pymc} \citep{AbrilPla2023PyMCAM}, a Python probabilistic programming library.
\texttt{pymc} automatically assigns a Markov Chain Monte Carlo (MCMC) sampler to perform inference; by default \texttt{pymc} uses a Hamiltonian Monte Carlo sampler \citep{hoffman2014nuts}. 
Crucially, by using a probabilistic program, we decouple modeling from inference: the proposal LM's role is to build a good model, while \texttt{pymc} takes care of inference.
Our approach of offloading computations to an external tool is connected to recent work enhancing LMs by giving them tools (\eg Python interpreter) \citep{gal_pal, Schick2023ToolformerLM}.

\paragraphspace{}
\paragraph{Model Criticism Step}
In the \textit{criticism step}, we ask the critic LM, $p_{LM}$, to produce natural language criticism of fitted models; we use this criticism to drive model revision. 
First, we can obtain a scalar score measuring the model fit (\eg ELPD LOO) for each proposed model.
Second, to enable the critic LM to do something akin to a posterior predictive check, we obtain samples from the posterior predictive distribution (\eg $p(x'|z, x) = \int_z p(x' | z) p (z|x)$) and then compute summary statistics of these posterior predictive samples \citep{Gelman2013BayesianDA}.
For simplicity, we compute the posterior predictive means and variances for each fitted probabilistic program.
We then provide $p_{LM}$ 
with select probabilistic programs, their scores $\mathcal{S}$ (\eg ELPD LOO), the posterior predictive means and variances $\mathcal{P}$, and the dataset itself $\mathcal{D}$; we explore both visual and textual dataframe based representations of the posterior predictive and the dataset.
Finally, we ask {$p_{LM}$} 
to distill this criticism in natural language (\eg von Bertalanffy growth function with informative priors) which we use to drive the next round of model building \citep{shinn2023reflexion}.

The key design choice is which programs to provide to {$p_{LM}$} 
(\eg \textit{critic exemplars}).
Naively, we could provide all proposed programs across all rounds to {$p_{LM}$}.
However, this list grows each round and has redundancy. Furthermore, LMs struggle to reason over long contexts.
We therefore explore a simple approach to selecting exemplars where we provide $p_{LM}$
with the top $d$ programs $\tilde{z}_1, \ldots \tilde{z}_d$ from the current round $t$.
To produce $h^{t+1}$, we sample: 
\begin{align*}
    h^{t+1} \sim p_{\text{LM}}(\cdot | \tilde{z}_1, \ldots, \tilde{z}_d, \mathcal{D}, \mathcal{S}, \mathcal{P}).    
\end{align*}
In the appendix, we report additional results for an approach inspired by a state-space update \citep{baum1966statistical, kalman1960new, rabiner89}.
To avoid storing the entire history of proposed programs, we interpret {$h^{t+1}$} as a latent state and compute it using the previous state {$h^{t}$} and the new fitted programs $\{z_i\}_{i=1}^m$ at round $t$: 
\begin{align*}
    h^{t+1} \sim p_{\text{LM}}(\cdot | {z_1}, \ldots, {z_m}, h^{t}, \mathcal{D}, \mathcal{S}, \mathcal{P}).    
\end{align*}
In practice, we implement this by asking $p_{\text{LM}}$ to add and delete hypotheses. 
We find that these two approaches have similar performances. 
We run this criticism step at a temperature of 0.0 to limit stochasticity in the criticism produced.

We refer to our full LM-driven automated Box's loop as \texttt{BoxLM}.

\section{Experiments}
\subsection{Searching over a DSL: automated Gaussian process kernel discovery}
\begin{figure*}[htpb]
\begin{minipage}
{0.40\textwidth}
\begin{tabular}{lrrrrrr}
\toprule
\text{Dataset} & \text{BoxLM+} & \text{BoxLM} & \text{Periodic} & \text{AS} & \text{SM} & \text{N-BEATS} \\
\midrule
\text{Air} & 0.08 & \underline{0.07} & 0.15 & 0.19 & \textbf{0.06} & 0.22 \\
\text{Beer} & 0.07 & 0.22 & 0.15 & \underline{0.06} & \textbf{0.05} & 0.02 \\
\text{Heart} & \underline{0.21} & \underline{0.21} & \textbf{0.20} & \underline{0.21} & \underline{0.21} & 0.07 \\
\text{Milk} & 0.12 & \underline{0.10} & \underline{0.10} & 0.11 & \textbf{0.09} & 0.04 \\
\text{Wine} & \underline{0.14} & 0.16 & 0.21 & \textbf{0.13} & 0.18 & 0.17 \\
\text{Wool} & 0.20 & \underline{0.19} & 0.19 & 0.23 & \textbf{0.13} & 0.18 \\
\bottomrule
\end{tabular}
\end{minipage}
\hspace{40mm}
\begin{minipage}{0.5\textwidth}
\includegraphics[scale=0.5]{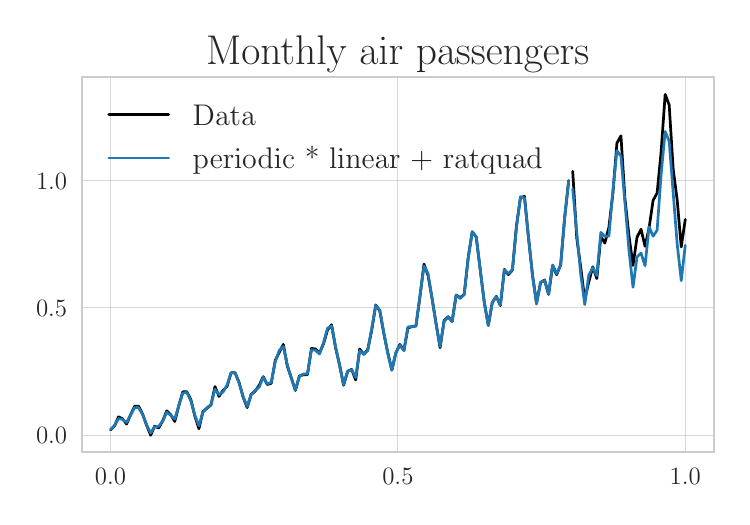}
    \end{minipage}   
\caption{
    \textbf{Test set performance on time series datasets.}
    {Our \texttt{BoxLM}} system identifies compositional kernels with performance on par with 
    strong baselines. 
    \textbf{(left)} 
    Comparison of \texttt{BoxLM} test mean absolute error (MAE) against Automatic Statistician  using greedy search (\texttt{AS}),
    spectral mixture kernel (\texttt{SM}), periodic kernel (\texttt{Periodic}), and \texttt{N-BEATS}.
    \texttt{BoxLM+} searches over an augmented  kernel space. 
    We bold the best and underline the second best among the GP methods, treating \texttt{N-BEATS} as a powerful non-GP-constrained baseline.  
    \textbf{(right)} Extrapolations from GP with a \texttt{BoxLM}-discovered kernel.
    }
\vspace{2mm}
\label{fig:gp_extrapolations}
\end{figure*}

We first evaluate LM's ability to search over a constrained space of models; in some settings, a domain expert may require the modeler to search over a DSL for reasons such as interpretability. 
In particular, we consider time-series modeling with Gaussian processes (GPs) as in the Automatic 
Statistician \citep{pmlr-v28-duvenaud13}.
GPs are probabilistic models that specify a distribution over functions.
GPs are defined by a mean function and a positive-definite kernel function $k(\mathbf{x},\ \mathbf{x'})$ that gives the covariance between~$f(\mathbf{x})$ and~$f(\mathbf{x'})$ as a function of~$\mathbf{x}$ and~$\mathbf{x'}$.
The properties of a GP depend significantly on the kernel and therefore choosing a kernel that reflects domain knowledge (\eg linearity, periodicity) is a key design choice.
One common way to produce a more flexible kernel is to compose kernels via addition and multiplication, leveraging the closure properties of kernels.
Here, we evaluate LMs' ability to search over a space of kernels to identify an appropriate composition of kernels. 
\paragraph{Setup}
Mirroring \citet{pmlr-v28-duvenaud13}, we define a space of base kernels. 
In the prompt, we ask $q_{LM}$ 
to perform one of the three operations (addition, multiplication, and replacement) on one of the in-context exemplar programs.   
For the first round, we ask $q_{LM}$ 
to produce an initial guess based on the structure of the dataset, which we provide as a plot.
Given a kernel expression, we learn the kernel hyperparameters via gradient-based optimization of the marginal likelihood.
\paragraphspace{}
\paragraph{Results}
We evaluate our method on six common univariate time series datasets.
We compare against the Automatic
Statistician, a greedy algorithm proposed by \citet{pmlr-v28-duvenaud13} (run until a depth of 10), and two GP baselines: a periodic kernel and a spectral mixture kernel, a strong non-compositional baseline. 
We also compare against \texttt{N-BEATS}, a strong neural baseline \citep{wilson2013kernels, Oreshkin2019NBEATSNB}. 
In Figure~\ref{fig:gp_extrapolations}, we compare the mean absolute error (MAE) on held-out test data for all datasets. 
To account for stochasticity in the Automatic  
Statistician implementation we used \citep{Saad2023SequentialMC} and stochasticity in different repetitions of our pipeline, we average the test MAE across three model runs. 
Our method, denoted \texttt{LM} in the table, matches the performance of the Automatic 
Statistician, showing that \texttt{BoxLM}
can efficiently search over a constrained space of models. 
We also experiment with augmenting the base kernel space with additional kernels (denoted {\texttt{BoxLM+}} in the table). 
In some cases, this additional flexibility is beneficial.
For example, by using the additional kernels, {\texttt{BoxLM+}} can much better capture the Australian beer sales dataset than \texttt{LM}; in other cases, the additional flexibility does not appreciably improve performance.
On the right panel, we show the extrapolations and identified kernel for the monthly air passenger dataset; {\texttt{BoxLM+}} identifies a kernel with a periodic times linear component to capture the increasing amplitude in the data.
\subsection{Open-ended probabilistic model discovery for real-world datasets}
By integrating LMs into the model discovery process, we can search over a much broader class of models. 
Here, we explore \texttt{BoxLMs}' ability to automatically construct \texttt{pymc} probabilistic programs \citep{AbrilPla2023PyMCAM} for datasets.
\vspace{-3mm}
\paragraph{Dataset} 
We consider four real world datasets from the Stan PosteriorDB dataset \citep{Magnusson_posteriordb_a_set_2023}: {(1)} a dataset consisting of average improvements in SAT scores after an SAT improvement program across eight different high schools, {(2)} a dataset consisting of ages of twenty-seven dugongs and their lengths, {(3)} a surgical dataset consisting of mortality rates in twelve hospitals performing cardiac surgery on babies, and {(4)} a dataset of peregrine population counts in the French Jura from 1964 to 2003.
Each dataset has an associated human expert written probabilistic program in Stan, which we translate into {\texttt{pymc}}.
These expert programs are generally open-source contributions from the Stan developer community.
The datasets cover common modeling motifs, such as hierarchical modeling and regression. 
\begin{figure*}[h]
    \centering
\includegraphics[scale=0.45]{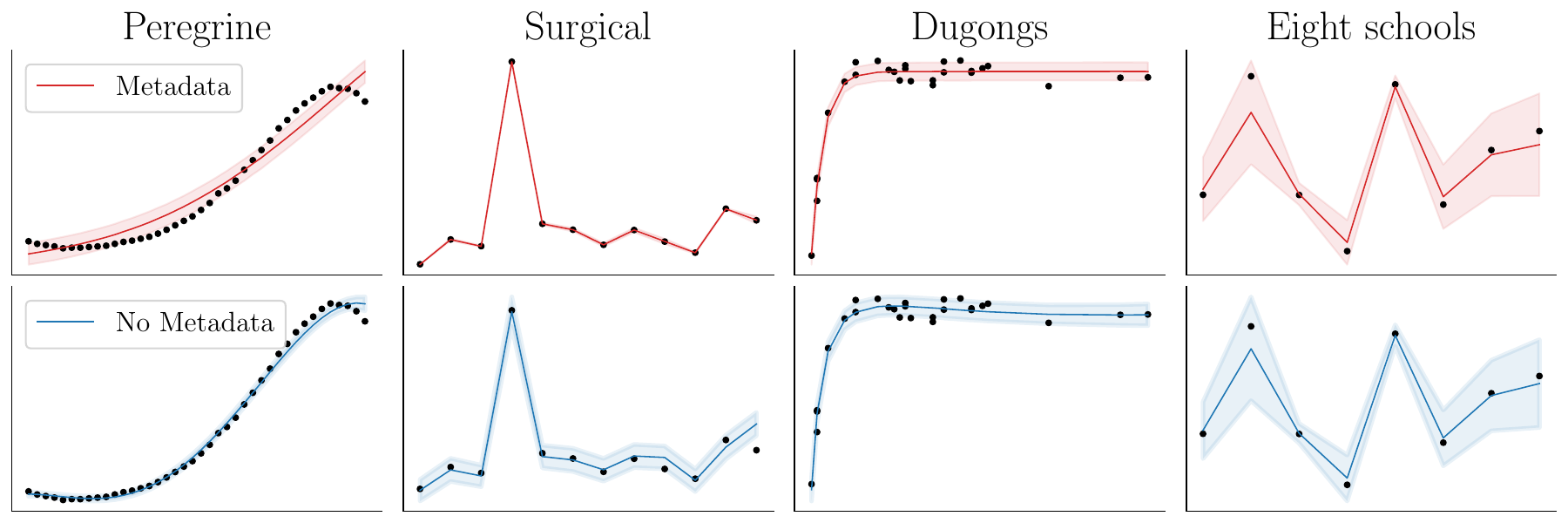}
\label{fig:enter-label}
\caption{\textbf{Domain knowledge shapes \texttt{BoxLM} modeling approaches.}
In the top row, we keep all metadata (\eg dataset description). In the bottom row, we remove metadata that reveals information about the domain. This leads to qualitatively different approaches for three datasets; for eight schools, \texttt{BoxLM} discovers a hierarchical model even without metadata.
We list the corresponding programs for these different ablations: 
\textbf{(top row)}
$\frac{K}{(1 + ((K - P_0) / P_0) \exp(-rt))}$,  
$\text{BetaBin}(n, \alpha, \beta)$, 
$L_{\text{inf}}(1 - \exp(-k (\text{age} - t_0)))$;
\textbf{(bottom row)}
$a + b x_0 + c x_0^2 + dx_0^3 + e x_0^4$, 
$\alpha + \beta x_0 + \gamma x_0^2$,
$\alpha + \beta_1\log x_0 + \beta_2 \log x_0^2 + \beta_3 \log x_0^3$.} 
\label{fig:anomalies}
\end{figure*}

\paragraphspace{}
\paragraph{Ablations} 
To study how domain knowledge affects the LM's modeling capabilities and to mitigate concerns about dataset leakage,  we consider two ablations. 
In the first ablation, we construct a \textit{simulated} analog for each dataset: for each dataset, we generate synthetic observations by sampling from the prior of a generative model or by sampling from a model with fixed parameters. 
We denote these datasets as \textit{simulated} datasets.
For example, to generate a simulated dataset for the \texttt{eight schools} dataset, we sample from the prior of the human expert program. 
In some cases, such as the \texttt{dugongs} dataset, the prior distributions over parameters are highly unconstrained and we therefore fix values for the parameters instead.
For example, to generate a simulated analog of the \texttt{dugongs} dataset, we fix the values of the parameters $\alpha, \beta, \gamma$ used in the expert model $y_i = \alpha - \beta \gamma^{x_i} + \epsilon_i$ where $\epsilon_i \sim \mathcal{N}(0, \sigma^2)$. 
Since it is unlikely that these simulated datasets appeared in the training data, this ablation helps mitigate concerns about dataset leakage.

In the second ablation, we remove the dataset \textit{metadata} from the LM prompt. 
In particular, we remove the dataset description, replace the column names with domain-agnostic column names (\eg $x_0, y$), and replace the axis labels with uninformative labels.
We denote these datasets as \textit{no metadata} datasets.
By removing this metadata, we can characterize how LMs use domain knowledge.
\paragraphspace{}
\paragraph{Quantitative Results}
In Table~\ref{tab:stan_ablations}, we compare the performance of our method across different datasets and ablations relative to the expert programs. 
We emphasize that the expert programs are strong baselines, especially for the simulated datasets where the expert programs generated the data. 
We highlight numbers corresponding to significant differences, where significance is defined as an ELPD LOO difference of greater than four times the standard error estimate. 

\texttt{BoxLM} reliably identifies programs on par with expert programs; we validate convergence using standard Markov Chain Monte Carlo diagnostics. 
Interestingly,  
removing metadata or switching to simulated datasets does not generally reduce performance relative to the expert program. 
The main exception is the \texttt{surgical} dataset. 
By replacing the labels with $x_0$ and $y$,  \texttt{BoxLM} formulates this problem as a regression problem instead of using a hierarchical model like the expert model.
For the \texttt{peregrine} dataset, removing metadata improves performance.
We discuss this further in the next section.
\paragraph{Qualitative Analysis: Language Models are Domain Experts}
Earlier, we asked whether LMs can reliably play the role of a domain expert. 
In our ablations, we study how metadata influences the model search process by examining how removing metadata changes the programs proposed by \texttt{BoxLM}. In Figure~\ref{fig:anomalies}, we plot the model fit and list the corresponding {\texttt{BoxLM}} programs in the figure caption.
For the \texttt{peregrine} dataset, when given all the metadata, {\texttt{BoxLM}} models this dataset using a logistic growth model, which is motivated by the problem domain.
However, logistic growth models cannot capture the initial decline in population. 
Interestingly, when we remove all metadata,  {\texttt{BoxLM}} performs better, because it uses a quartic polynomial to model the data. 
This closes the gap in performance with the expert program.
This highlights how prior knowledge can have a (sometimes overly) strong influence on the modeling choices of {\texttt{BoxLM}}.
We observe a similar trend with the \texttt{dugongs} growth curve dataset. 
When given metadata, {\texttt{BoxLM}} uses a von Bertalanffy growth function \citep{Bertalanffy1949ProblemsOO}, which is commonly used to model animal growth. 
When we remove all metadata, {\texttt{BoxLM}} uses a polynomial function with a logarithmically transformed value for the inputs. Here, both modeling approaches fit the data well.
Interestingly, {\texttt{BoxLM}} sets the prior parameters in these models in a data-informed way. For example, {\texttt{BoxLM}} sets the asymptotic length in the von Bertanlanffy function based on the observed lengths in the dataset. See code snippet in Figure~\ref{code:corruption_compared_1} of the Appendix.
For the eight schools dataset, {\texttt{BoxLM}} identifies a hierarchical model even without metadata, illustrating that model criticism can compensate for lack of domain knowledge.
We illustrate the improvement round to round in Figure~\ref{fig:improvement}.
\begin{table}[ht]
\centering
\begin{tabular}{lll}
\toprule
 Dataset & Expert & LM       \\
\midrule
Eight schools & \underline{-30.70} & 
\underline{-30.42} \\
Eight schools sim & 
\underline{-18.09} & \underline{-18.31} \\
Eight schools sim no metadata & \underline{-18.09} & \underline{-16.36} \\
Dugongs & 
\underline{22.52} &
\underline{23.40} \\
Dugongs sim & 
 \underline{50.04} & 
 \underline{57.40} \\
Dugongs sim no metadata & \underline{50.04} & \underline{55.24} \\
Surgical & \underline{-40.29} & -38.03 \\
Surgical sim & \underline{-39.80} & \underline{-38.38} \\
Surgical sim no metadata & \textbf{-39.80} & -63.72 \\
Peregrine & \textbf{-142.19} & -173.11 \\
Peregrine sim & \textbf{-130.48} & -179.06 \\
Peregrine sim no meta & \underline{-130.48} & \underline{-136.39} \\
\bottomrule
\end{tabular}
\caption{\textbf{Comparison of \texttt{BoxLM} programs against expert programs} We perform this comparison across four different datasets and two different ablations that replace observations with synthetic observations and remove all metadata. 
We report the expected predictive log density estimated via leave-one-out cross validation.
        We bold statistically significant differences and underline non-significant differences.     
        LM programs match the performance of human expert programs on 9/12 datasets. 
}
\label{tab:stan_ablations}
\end{table}

\subsection{Improving classic models under modeling constraints}
\label{sec:lotka_volterra}
\begin{figure*}[htpb] 
\centering
\includegraphics[width=0.45\textwidth]{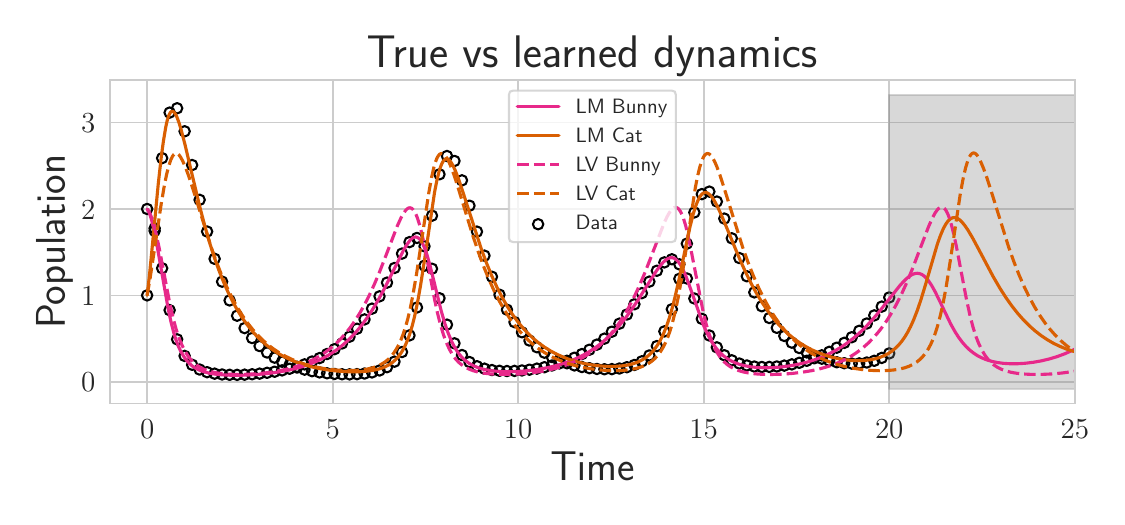} 
\includegraphics[width=0.52\textwidth]{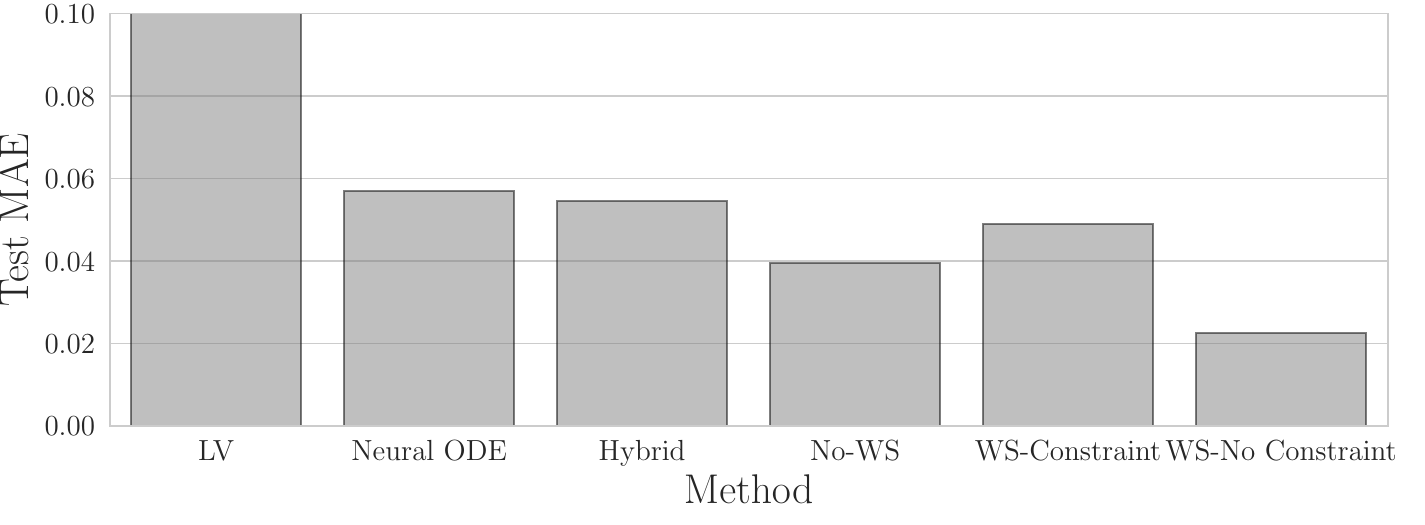} 
\caption{\textbf{Correcting misspecified Lotka-Volterra dynamics.} {\texttt{BoxLM}} can introduce corrections to standard Lotka-Volterra dynamics (no warm-start) 
and a hybrid neural {ODE} approach (warm-start) 
that outperform several baselines. 
\textbf{(left)} {\texttt{LM}}-proposed model predictions on training data and extrapolations (grey region). \textbf{(right)} 
Test MAE of \texttt{LM} models (\texttt{No-WS}, \texttt{WS-Constraint}, and \texttt{WS-No Constraint}) compared to
the standard Lotka-Volterra model {\texttt{LV}}, a \texttt{Neural ODE}, and
a hybrid Neural ODE model with a multiplicative correction to the prey-predator dynamics ({\texttt{Hybrid}}).
}
\label{fig:lotka_volterra}
\end{figure*}

\begin{figure*}[ht!]
\centering
\begin{tcolorbox}[
title={LM proposed Lotka-Volterra programs}]
\fontsize{7pt}{7pt}\selectfont
\ttfamily
\begin{lstlisting}[language=Python]
def no_warm_start(t, y, _coeffs):
  # Hypothesis 1: Logistic growth for prey with carrying capacity kappa
  # Hypothesis 2: Saturation effect in predation rate with parameter psi
  db_dt = alpha * b * (1 - b / kappa) - beta * b * c / (1 + psi * b)
  dc_dt = -gamma * c + delta * b * c - epsilon * c**2
\end{lstlisting}

\begin{lstlisting}[language=Python]
def warm_start_constrained(t, y, _mlp):
  inputs = jnp.array([b])  # Input is prey density (bunny population)
  # Neural network learns a modifier for the handling time based on prey density
  handling_time_modifier = _mlp(inputs)
  # Modulate the predation rate beta in a non-linear manner according to handling time
  db_dt = alpha * b - beta * b * c / (1 + handling_time_modifier[0] * b)
  dc_dt = -gamma * c + delta * b * c / (1 + handling_time_modifier[0] * b)
\end{lstlisting}

\begin{lstlisting}[language=Python]
def warm_start_unconstrained(t, y, _mlp):
  inputs = jfu.ravel_pytree((b, c))[0]
  mlp_output = _mlp(inputs)
  # Fine-tuned scaling of the MLP output to match the amplitude of data more closely
  db_dt = alpha * b - beta * b * c + 0.02 * mlp_output[0] # Reduced scaling for bunnies
  dc_dt = -gamma * c + delta * b * c + 0.06 * mlp_output[1] # Increased scaling for cats
\end{lstlisting}

\end{tcolorbox}
\caption{\textbf{\texttt{BoxLM} can propose corrections to ODEs.} \textbf{(top)} In the no warm-start {(\texttt{No-WS})} variation, {\texttt{BoxLM}} introduces corrections informed by domain knowledge of predator-prey models (carrying capacity, predation saturation).
\textbf{(middle)} 
When prompted to introduce neural networks in an interpretable way (\texttt{WS-Constraint}), one strategy {\texttt{BoxLM}} proposes is to make the handling time parameter depend non-linearly on the prey density, extending a traditional approach to modeling predation saturation. 
\textbf{(bottom)} 
When prompted to introduce neural networks without constraints (\texttt{WS-No Constraint}), {\texttt{BoxLM}} introduces additive MLP-parameterized corrections and adjusts the scaling factors.}
\vspace{-3mm}
\label{code:lotka-volterra_main}
\end{figure*}

In the previous experiment, we explored {\texttt{BoxLM}}'s ability to identify models \textit{tabula rasa} (\eg without any initial seed model).
However, in many scientific settings, we begin with a well-known model and are tasked with improving it. 
Here, we explore {\texttt{BoxLM}}'s ability to improve upon a Lotka-Volterra model of predator-prey dynamics.
In addition, a crucial component of model discovery is respecting ``soft'' modeling constraints that are easy to express in natural language but hard to formalize (\eg ecologists should think this is a plausible model). 
We therefore illustrate another advantage of \texttt{BoxLM}: we can express these modeling constraints in natural language and use them to drive LM proposals.
\paragraphspace{}
\paragraph{Dataset}
To create our dataset, we simulate data from the following ``perturbed'' Lotka-Volterra dynamics
\begin{align*}
\frac{db}{dt} &= \alpha b - \beta b c && \\
\frac{dc}{dt} &= -\gamma c + \delta bc^{0.95}.
\label{eq:perturbed_lv}
\end{align*}
\paragraph{Setup}
{\texttt{BoxLM}} is tasked with implementing an ODE model in Python using \texttt{Jax} \citep{jax2018github}. 
Estimating the parameters of Lotka-Volterra models via Bayesian inference is challenging.
We instead learn the parameters via gradient descent which can be straightforwardly implemented using modern automatic differentiation libraries \citep{chen_neural_ode,kidger2021on}; in particular, we use \texttt{diffrax} \citep{kidger2021on}, a \texttt{Jax}-based ODE library that supports learning ODE parameters via backpropagation. 
We consider three variations: \textit{warm-start with constraints} {(WS-Constraint)}, \textit{warm-start with no constraints} {(WS-No Constraint)}, and \textit{no warm-start} {(No-WS)} that differ in their initial seed program and the initial instructions which express modeling constraints.
In all variations, we provide the LM with a scatter plot of training datapoints. 
In the \textit{no warm-start} variation, we provide the LM with an implementation of standard Lotka-Volterra dynamics using \texttt{diffrax} and the predictions obtained from fitting standard Lotka-Volterra to the training data. 
In both \textit{warm-start} variations, we provide the LM with {(1)} an implementation of a hybrid neural ODE that introduces an additive correction term to the predator dynamics, parameterized by a multilayer perceptron (MLP), and {(2)} the predictions obtained from fitting this model to the training data. 
In the constrained warm-start variation, we ask the LM to produce a hybrid neural ODE model that is interpretable to an ecology expert who suggested a Holling's type II response \citep{rosenzweig1963graphical}.
In the unconstrained warm-start variation, we provide the same seed program but do not impose this additional {interpretability} constraint. 
For models with both neural and physical components, we employ a two-stage learning procedure so that the neural component does not dominate the dynamics; see Appendix~\ref{sec:appendix_lotka_volterra} for details.
\vspace{-5mm}
\paragraph{Results}
In Figure~\ref{fig:lotka_volterra} (left), we plot the predictions obtained from integrating the learned dynamics for an ODE proposed by {\texttt{BoxLM}}, the training data points generated from the true dynamics, and the predictions from the standard Lotka-Volterra model. 
We fit free parameters to the training data via gradient descent.
The grey region indicates the extrapolation region. 

{\texttt{BoxLM}} can significantly improve upon the standard Lotka-Volterra model by introducing corrections to the dynamics (Figure~\ref{fig:lotka_volterra}).
The standard Lotka-Volterra model cannot capture the decreasing amplitude in the data; furthermore, there is a slight phase shift relative to the training datapoints. 
In contrast, {\texttt{BoxLM}} identifies an ODE that captures these properties and extrapolates accurately. 
In Figure~\ref{fig:lotka_volterra} (right), we compare these programs against a neural ODE baseline, and a hybrid neural ODE baseline that incorporates a multiplicative correction (parameterized by an MLP) to the predator-prey interaction term in the predator equation. 
See Section~\ref{sec:appendix_lotka_volterra} of the Appendix for details on these baselines. 
For the {\texttt{LM}} variations, we report the average test MAE across three runs.
In Figure~\ref{fig:lotka_volterra}, we see that \textit{all} {\texttt{BoxLM}} variations outperform the baselines. 

In Figure~\ref{code:lotka-volterra_main}, we present code snippets corresponding to representative programs proposed in the constrained and unconstrained variations.
These snippets show how natural language constraints can guide {\texttt{BoxLM}} towards more flexible models that retain interpretability.  
For the variation with no natural language constraints (WS-No Constraint), {\texttt{BoxLM}} takes a purely empirical approach. 
In particular, {\texttt{BoxLM}} adjusts the scaling terms on the additive {\texttt{MLP}} correction term. 
For the WS-Constraint variation,
{\texttt{BoxLM}} proposes a hybrid approach integrating the neural approach in the prompt with classic models in the literature; importantly, even though {\texttt{BoxLM}} is asked to balance interpretability and flexibility, {\texttt{BoxLM}} still identifies programs that outperform the neural ODE and standard Lotka-Volterra baselines.  
One approach {\texttt{BoxLM}} proposes is an extension of the Rosenzweig and MacArthur model with a Holling's type II functional response \citep{rosenzweig1963graphical} to allow a static parameter to depend dynamically on the prey density: {\texttt{BoxLM}} models the handling time, or the time a predator spends ``processing'' a prey, as a nonlinear function
of the prey density via an MLP.
These results show how we can use natural language to drive {\texttt{BoxLM}} towards models that balance flexibility and interpretability. 

\section{Conclusion}
We introduced a method for leveraging LMs for automated model discovery. 
Our method can identify models that perform favorably against strong baselines and improve upon expert models. 
We also studied how domain knowledge and natural language constraints influence our system. 
Altogether, our results highlight the compelling advantages of LM-driven statistical model discovery.

Our work has important limitations that motivate future research. 
First, we focused on modeling static datasets. An interesting direction could be leveraging LMs for active data collection.
Second, since our tasks were restricted to one-dimensional datasets, simple model criticism statistics were sufficient and therefore decided in advance (residuals, posterior predictive mean). 
Another interesting future direction could be fully automating the criticism step.
Finally, while in-context learning was effective in our tasks, we could explore finetuning techniques for training a language model to produce better probabilistic programs.
\begin{figure}[ht] 
\label{fig:lotka_volterra_1}
\centering
\includegraphics[width=0.275\textwidth]{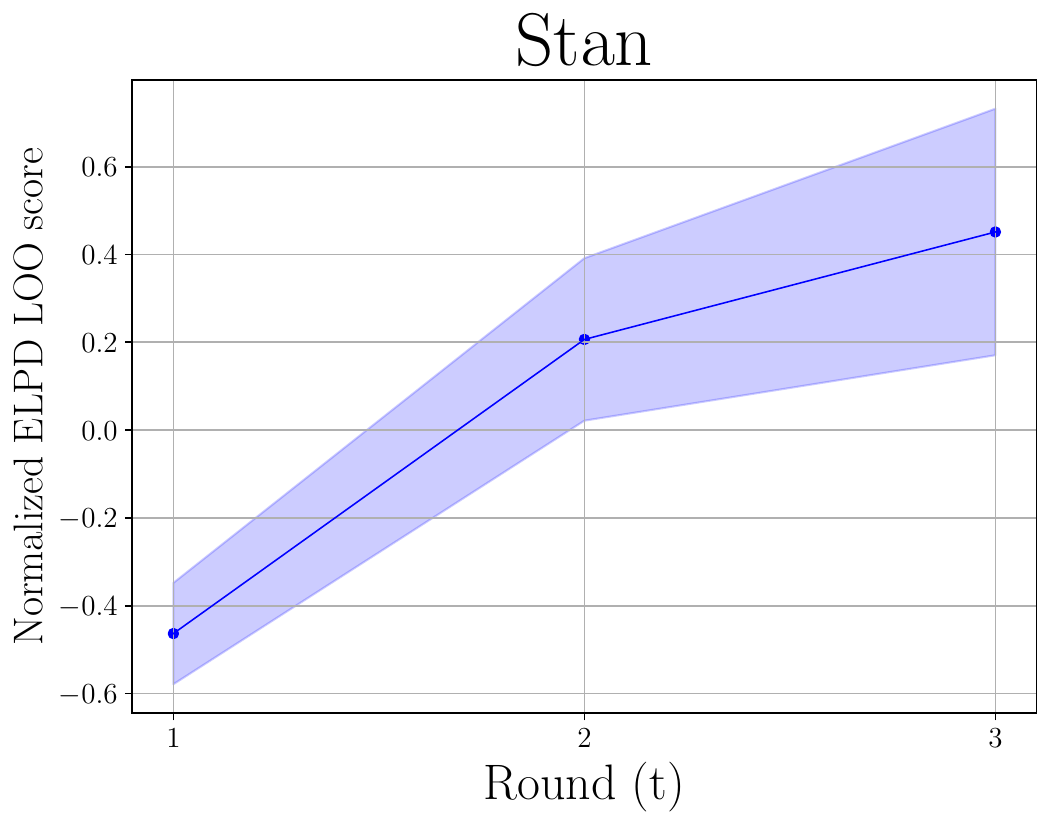} 
\includegraphics[width=0.275\textwidth]{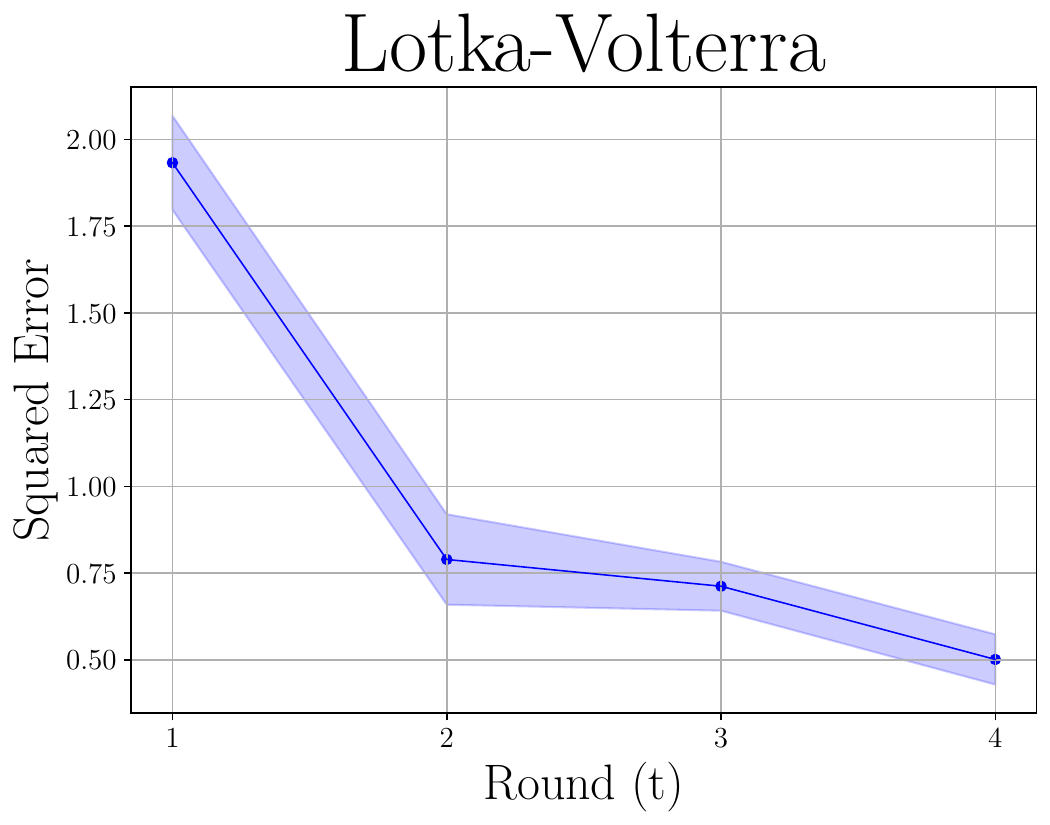} 
\caption{\textbf{Round-to-round improvement.} Model revision leads to improvements on average. (Improvement is not necessarily monotonic for a given dataset and run.)
\textbf{(top)} ELPD LOO score vs. round for Stan experiments. We normalize the ELPD LOO scores across programs proposed for 3 rounds of Box's loop and average across datasets for the no metadata condition. 
Larger is better and error bars correspond to standard error. 
\textbf{(bottom)} Squared error vs round for LV experiment. We report the squared error averaged across three different runs, for the warm-start, no constraint condition; smaller is better.
\label{fig:improvement}}
\end{figure}

\newpage
\section*{Acknowledgements}
This work was supported in part by AFOSR Grant FA9550-21-1-0397, ONR Grant N00014-22-1-2110, and an NSF Expeditions Grant, Award
Number (FAIN) 1918771.
EBF is a Chan Zuckerberg Biohub – San Francisco Investigator.

We thank Eric Zelikman and Neil Band for detailed feedback on this paper.
We also thank Omar Shaikh and Jensen Gao for helpful discussions.

\section*{Impact Statement}
This paper presents work whose goal is to partially automate statistical modeling. There are many potential societal consequences of our work, none which we feel must be specifically highlighted here.

\newpage
\bibliography{references}
\bibliographystyle{icml2024}

\newpage
\appendix
\onecolumn
\section{Gaussian Process experiments}
In the prompt, we ask the LM to use the following operations.
\begin{enumerate}
    \item Replace a subexpression $\mathcal{S}$ with $\mathcal{S}+\mathcal{B}$, where $\mathcal{B}$ is any base kernel.  
    \item Replace a subexpression $\mathcal{S}$ with $\mathcal{S} x \mathcal{B}$, where $\mathcal{B}$ is any base kernel. 
    \item Any base kernel $\mathcal{B}$ may be replaced with any other base kernel family.    
\end{enumerate}

\paragraph{LM hyperparameters}
We provide the LM with the following kernels: Exponentiated Quadratic, Periodic, Linear, and Polynomial. 
We run our pipeline for two rounds with three proposals each round. 
We use a temperature of 0.2 for the Proposal LM and temperature of 0.0 for the Critic LM.
We use three in-context exemplars. 
Our Critic LM conditions on the best twelve programs so far.

In the augmented variation, we also provide the LM with the following additional kernels: Matern32, Matern52, Cosine, and the Rational Quadratic kernel.
In the augumented variation, we run our pipeline for three rounds with eight proposals each round. 
We use a temperature of 0.7 for the Proposal LM and temperature of 0.0 for the Critic LM.
We use three in-context exemplars. 
Our Critic LM conditions on the best twelve programs so far. 

The marginal likelihood can be multimodal in the parameters of the periodic kernel.
Therefore, following \citet{pmlr-v28-duvenaud13}, if the proposed kernel has periodic components, we initialize the period at five different initial values, optimize the marginal likelihood starting from those different initializations, and choose the kernel hyperparameters with the highest marginal likelihood across those initializations. 

\paragraph{Spectral Mixture kernel}
We use a GP with a spectral mixture kernel \citep{wilson2013kernels} with 5 mixture components. 
For each dataset, we randomly initialize the parameters of the mixture and choose the kernel hyperparameters with the highest log marginal likelihood across five random initializations.

\section{Stan Experiments}

\paragraph{Eight Schools Dataset}
This dataset consists of eight observations: the estimated treatment effect of a SAT coaching program and the standard error of the treatment effect. 

\paragraph{Peregrine dataset}
This dataset consists of peregrine population counts in the French Jura from 1964 to 2003 (40 observations in total).

\paragraph{Dugongs Dataset}
The ages and lengths of 27 captured dugongs (sea cows).

\paragraph{Surgical Dataset}
The mortality rates in 12 hospitals performing cardiac surgery on babies.

\paragraph{LM hyperparameters}
We run our pipeline for three rounds with eight proposals each round. 
We use a temperature of 0.7 for the Proposal LM and temperature of 0.0 for the Critic LM.
We use three in-context exemplars. 
Our Critic LM conditions on the best twelve programs so far. 

\paragraph{Markov Chain Monte Carlo diagnostics}
We evaluated the fidelity of the learned posteriors using the Gelman-Rubin $\hat{R}$ diagnostic~\citep{gelman1992} and by examining the Bulk Effective Sample Size (ESS). 
In particular, the programs reported in the table all had $\hat{R} \leq 1.01$ and mean bulk ESS $>= 400$ per chain. 
\begin{figure*}
\centering
\begin{tcolorbox}[
title={Language model produced \texttt{pymc} code snippet for modeling Peregrine population count}]
\fontsize{7pt}{7pt}\selectfont
\ttfamily
\begin{lstlisting}[language=Python]
def dugongs_gen_model():
  # Priors for von Bertalanffy growth function parameters
  L_inf = pm.Uniform('L_inf', lower=0, upper=3) # Asymptotic length, bounded by observed data
  k = pm.Uniform('k', lower=0, upper=1) # Growth coefficient, bounded by reasonable values
  t0 = pm.Uniform('t0', lower=-5, upper=5) # Theoretical age at zero length
  # Expected length at each age using von Bertalanffy growth function
  expected_length = L_inf * (1 - pm.math.exp(-k * (ages - t0)))
  # Likelihood for observed lengths
  sigma = pm.Uniform('sigma', lower=0, upper=1) # Standard deviation of observed lengths around the mean
  y_obs = pm.Normal('y_obs', mu=expected_length, sigma=sigma, observed=lengths)
\end{lstlisting}
\end{tcolorbox}
\caption{\textbf{LM proposes programs informed by domain knowledge} LM chooses a model informed by the domain (animal length vs age) and sets the priors based on the dataset (\eg the largest length in the dataset is smaller than 3).
\label{code:corruption_compared_1}
}
\end{figure*}

\section{Lotka-Volterra}
\label{sec:appendix_lotka_volterra}
\paragraph{Dataset}
To create our dataset, we simulate data from the following ``perturbed'' Lotka-Volterra dynamics
\begin{align}
\frac{db}{dt} &= \alpha b - \beta b c && \\
\frac{dc}{dt} &= -\gamma c + \delta bc^{0.95}
\label{eq:perturbed_lv_appendix}
\end{align}
We set $\alpha = 0.9, \beta=1.1, \delta=-1.2, \gamma = 2.1$.
The parameter $\alpha$ characterizes the prey's maximum growth rate and 
$\beta$ controls how the predator population modulate the growth rate. 
The parameter $\gamma$ characterizes the prey's maximum death rate and 
$\delta$ controls how the predator's growth rate depends on the prey population density.
In contrast to the standard Lotka-Volterra dynamics, we raise $c$ to a fractional power.

We now describe the various baselines we compare against in Section~\ref{sec:lotka_volterra}.
\paragraph{Standard Lotka-Volterra}
We fit the free parameters of the standard Lotkva-Volterra differential equations. 
\begin{align}
\frac{db}{dt} &= \alpha b - \beta b c && \\
\frac{dc}{dt} &= -\gamma c + \delta bc
\label{eq:standard_lv}
\end{align}

\paragraph{Neural ODE baseline}
We parameterize the predator and prey equations with an MLP. 
We run a hyperparameter search over four widths (4, 8, 16, 32) and 3 depths (1,2,4).
We use a learning rate of 3e-3 and train using full-batch gradient descent with Adam for 1500 iterations.

\paragraph{Hybrid Neural ODE baseline}
We implement a Hybrid Neural ODE baseline that introduces a correction to the predator-prey interaction term. 
Note that, we follow a two-stage ``boosting'' type procedure to fit the parameters of the MLP.
First, we fit the free parameters $\alpha, \beta, \gamma, \delta$ to the data. 
We then \textit{freeze} those parameters and fit the MLP parameters.
Without this two-staged approach, 
the MLP term can dominate the dynamics.
The MLP term has one layer and four hidden units and we train the MLP with full batch gradient descent with Adam using a learning rate of 3e-3.
\begin{align}
\frac{db}{dt} &= \alpha \cdot b - \beta \cdot b \cdot c \\
\frac{dc}{dt} &= -\gamma \cdot c + \delta \cdot b \cdot (c + 0.1 \cdot \text{\text{mlp}}(b, c))
\label{eq:multiplicative_neural}
\end{align}

In the warm-start variations, we provide the LM with an initial hybrid Neural ODE baseline that introduces an additive correction to prey equation. 
The MLP term has one layer and four hidden units.
\begin{align}
\frac{db}{dt} &= \alpha \cdot b - \beta \cdot b \cdot c + 0.1 \cdot \text{\text{mlp}}(b, c) \\
\frac{dc}{dt} &= -\gamma \cdot c + \delta \cdot b \cdot c
\label{eq:additive_neural}
\end{align}

\paragraph{LM hyperparameters}
We run our pipeline for four rounds with twelve proposals each round. 
We use a temperature of 0.7 for the Proposal LM and temperature of 0.0 for the Critic LM.
We use three in-context exemplars. 
Our Critic LM conditions on the best twelve programs so far. 

\section{Failure rates of GPT-4 V proposed programs}
\begin{table}[htpb]
    \centering
    \begin{tabular}{ll}
        \hline
        \textbf{Model} & \textbf{ Percent successfully scored} \\
        \hline
        GPT-4 textual dataframe & 78\% \\
        \hline
        GPT-4 vision only & 70\% \\
        \hline
        GPT-3.5 & 76\% \\
        \hline
    \end{tabular}
    \caption{Percentage of \texttt{pymc} programs that we can successfully perform inference in for Stan experiments.}
    \label{tab:success_rate}
\end{table}

\section{Additional ODE results}
\begin{figure*}[ht!] 
\centering
\includegraphics[width=0.40\textwidth]{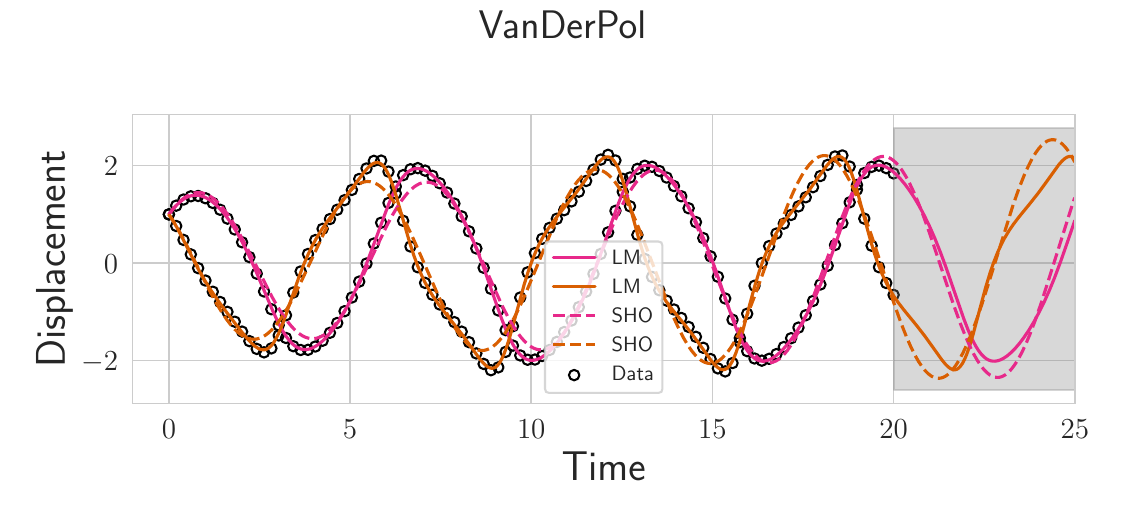} 
\includegraphics[width=0.40\textwidth]{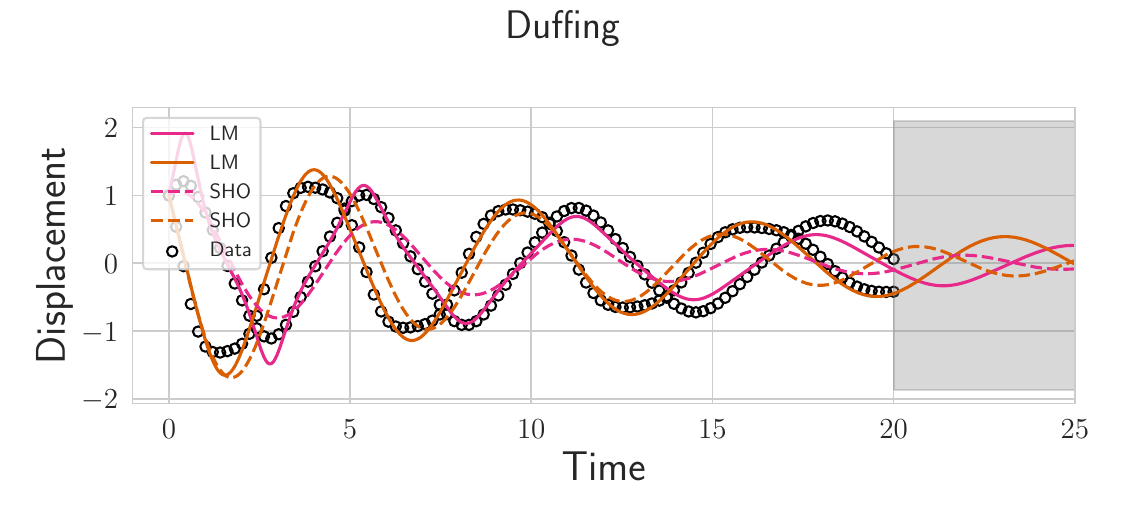} 
\includegraphics[width=0.40\textwidth]{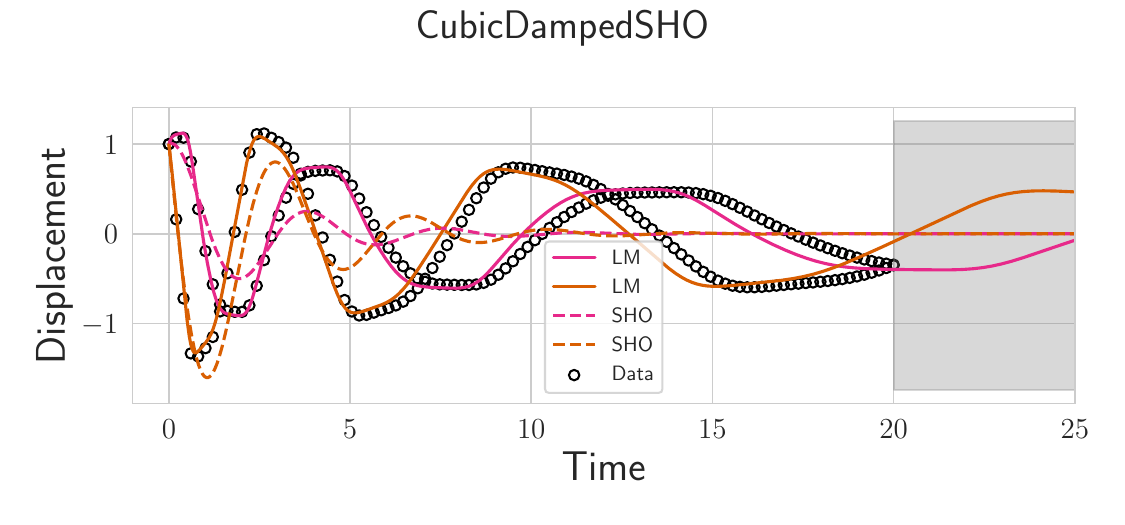} 
\caption{\textbf{Modeling nonlinear ODEs.} LM can introduce polynomial corrections to simple harmonic oscillator (SHO) that provide a better fit. Grey region indicates extrapolation region.
}
\label{fig:ode_additional}
\end{figure*}
We evaluated on three additional ODEs (nonlinear oscillators) from this github repository \href{https://github.com/dynamicslab/pysindy/blob/master/pysindy/utils/odes.py}{\texttt{pysindy}}.
These ODEs include a Duffing ODE, Van Der Pol oscillator, and a Cubic Harmonic Damped oscillator. 
We give \texttt{BoxLM} a simple harmonic oscillator to start and ask it to introduce polynomial corrections to improve the fit. 
Consistent with Section~\ref{sec:lotka_volterra}, our approach can generally improve upon a baseline ODE and extrapolate accurately into the test region. 

\section{Visual Interface/GPT-3.5 Ablations}
Context length limits are a potential limitation if we provide the dataset in textual format in the prompt. 
We therefore experiment with a visual-only variation. We remove all textual representations of datasets and model criticism statistics from the prompt and only provide visual plots of these datasets and statistics. 
We show this visual-only variation does not harm the performance relative to textual variation and should not suffer as much from context length limits as the dataset grows larger. 
We also show we can obtain similar results with GPT-3.5 Turbo, which is significantly less costly than GPT-4. 
\begin{table}[h!]
\centering
\begin{tabular}{llllll}
\toprule
Dataset & GPT-4 Text & GPT-4 Visual Only & GPT-3.5  & Expert \\
\midrule
Eight schools & -30.17  &  -30.40  & -30.44 & -30.70  \\
Dugongs & 22.61 & 23.76  & 21.01  & 22.52 \\
Surgical & -37.36 & -38.54 & -42.2  & -40.29 \\
Peregrine & -164.69 & -143.45 & -161.14  & -142.19 \\
\bottomrule
\end{tabular}
\caption{\textbf{Vision interface and model type ablations.} Comparison of GPT-4 with textual representation of data and model criticism statistics in prompt, GPT-4 with only visual representations (GPT-4 Visual Only) of data and model criticism statistics (\eg only plots), and GPT 3.5 Turbo against expert programs. The visual-only variation (GPT-4 Visual-Only) performs comparably to the textual variation (GPT-4 Text )and outperforms the textual variation on the Peregrine dataset. GPT-3.5 performs slightly worse on some datasets but comparably on most. We report the expected predictive log density (LOO). 
}
\label{tab:gpt_vision}
\end{table}

\section{State-space model hidden state update experiments}
We present the same results from the main text but take a state-space update inspired approach to computing the natural language criticism $h^t$. 
\begin{figure*}[ht]
\centering
\begin{tabular}{lrrrrrr}
\toprule
\text{Dataset} & \text{BoxLM+} State Space & \text{BoxLM} State Space & \text{Periodic} & \text{AS} & \text{SM} & \text{N-BEATS} \\
\midrule
\text{Air} & \textbf{0.04} & 0.34 & 0.15 & 0.19 & \underline{0.06} & 0.22 \\

\text{Beer} & 0.07 & 0.41 & 0.15 & \underline{0.06} & \textbf{0.05} & 0.02 \\

\text{Heart} &  \textbf{0.20} & 0.25 & \textbf{0.20} & \underline{0.21} & \underline{0.21} & 0.07 \\

\text{Milk} & \textbf{0.06} & 0.11 & \underline{0.10} & 0.11 & \underline{0.09} & 0.04 \\

\text{Wine} & \underline{0.17} & \underline{0.17} & 0.21 & \textbf{0.13} & 0.18 & 0.17 \\

\text{Wool} & 0.24 & \underline{0.15} & 0.19 & 0.23 & \textbf{0.13} & 0.18 \\

\bottomrule
\end{tabular}
\caption{
\textbf{Test set performance of \texttt{BoxLM} with state-space updated on time series datasets.}
Performance of {\texttt{BoxLM}} system with state space update for model criticism.
Comparison of \texttt{BoxLM} test mean absolute error (MAE) against Automatic Statistician  using greedy search (\texttt{AS}),
spectral mixture kernel (\texttt{SM}), periodic kernel (\texttt{Periodic}), and \texttt{N-BEATS}.
\texttt{BoxLM+} searches over an augmented  kernel space. 
We bold the best and underline the second best among the GP methods.
}
\label{fig:gp_extrapolations_ssm}
\vspace{-3mm}
\end{figure*}

\begin{table}[ht]
\centering
\begin{tabular}{lll}
\toprule
 Dataset & Expert & LM       \\
\midrule
Eight schools & \underline{-30.70} & \underline{-30.17} 
\\
Eight schools sim & 
\underline{-18.09} &  \underline{-18.39} \\
Eight schools sim no metadata & \underline{-18.09} & \underline{-18.90} \\
Dugongs & \underline{22.52} & \underline{22.61}\\
Dugongs sim & \underline{50.04} &  \underline{57.4} \\
Dugongs sim no metadata & 
\textbf{50.04} & 26.68 \\
Surgical & \textbf{-40.29} & -37.36 \\
Surgical sim & \underline{-39.80} & \underline{-38.45} \\
Surgical sim no metadata & \textbf{-39.80} & -58.72 \\
Peregrine & \textbf{-142.19} &   -164.69\\
Peregrine sim & \textbf{-130.48} & -177.15 \\
Peregrine sim no meta & \underline{-130.48} & \underline{-127.32} \\
\bottomrule
\end{tabular}
\caption{\textbf{Comparison of BoxLM with state-space update programs against expert programs} We perform this comparison across four different datasets and two different ablations that replace observations with synthetic observations and remove all metadata. 
We report the expected predictive log density estimated via leave-one-out cross validation.
We bold statistically significant differences and underline non-significant differences.     
}
\label{tab:stan_ablations_ssm}
\end{table}

\begin{figure*}[htpb] 
\centering
\includegraphics[width=0.75\textwidth]{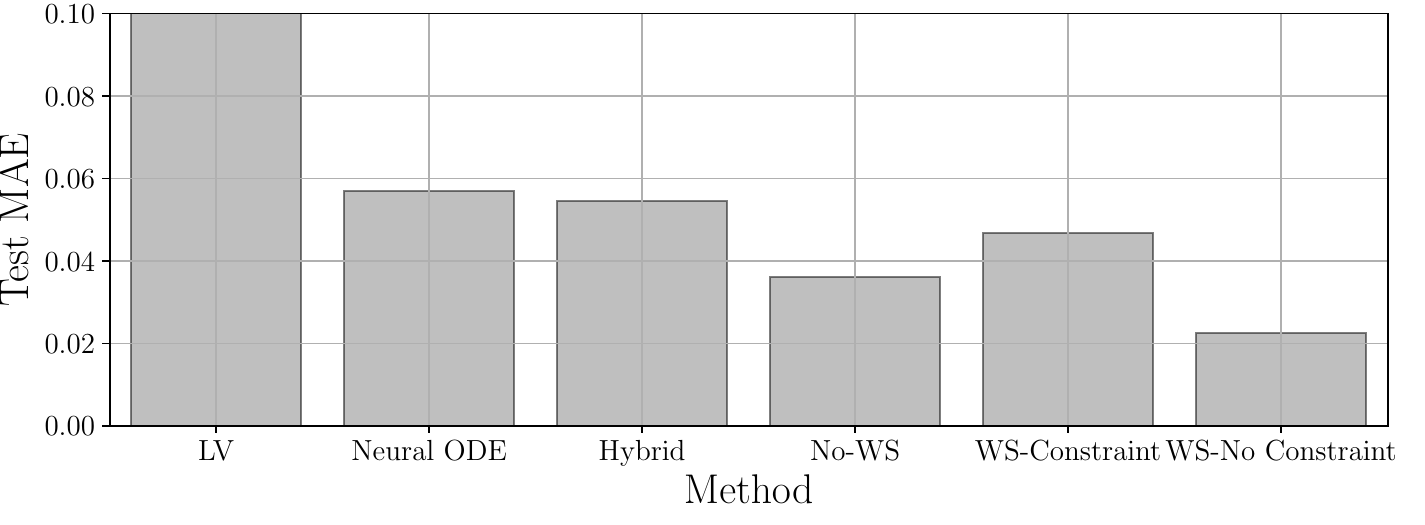} 
\caption{\textbf{Correcting misspecified Lotka-Volterra dynamics (\texttt{BoxLM} with state-space updates).} {\texttt{BoxLM}} can introduce corrections to standard Lotka-Volterra dynamics (no warm-start) 
and a hybrid neural {ODE} approach (warm-start) 
that outperform several baselines. 
Test MAE of \texttt{LM} models (\texttt{No-WS}, \texttt{WS-Constraint}, and \texttt{WS-No Constraint}) compared to
the standard Lotka-Volterra model {\texttt{LV}}, a \texttt{Neural ODE}, and
a hybrid Neural ODE model with a multiplicative correction to the prey-predator dynamics ({\texttt{Hybrid}}).
}
\label{fig:lotka_volterra_ssm}
\end{figure*}


\end{document}